\documentclass[]{acmart}

\AtBeginDocument{%
  \providecommand\BibTeX{{%
    \normalfont B\kern-0.5em{\scshape i\kern-0.25em b}\kern-0.8em\TeX}}}

\setcopyright{acmcopyright}
\copyrightyear{2020}
\acmYear{2020}
\acmDOI{10.1145/1122445.1122456}

\acmConference[ACM Transactions on Data Science]{ACM Transactions on Data Science}{March}{2020}

\begin{document}

\title[Multi-User Remote lab]{Multi-User Remote lab: Timetable Scheduling Using Simplex Nondominated Sorting Genetic Algorithm}

\author{Seid Miad Zandavi}
\affiliation{%
  \institution{School of Computer Science, The University of Sydney}
  \streetaddress{Camperdown}
  \city{Sydney}
  \country{Australia}}
  \email{miad.zandavi@sydney.edu.au}

\author{Vera Chung}
\affiliation{%
  \institution{School of Computer Science, The University of Sydney}
  \streetaddress{Camperdown}
  \city{Sydney}
  \country{Australia}}
\email{vera.chung@sydney.edu.au}

\author{Ali Anaissi}
\affiliation{%
  \institution{School of Computer Science, The University of Sydney}
  \city{Sydney}
  \country{Australia}}
  \email{ali.anaissi@sydney.edu.au}


\begin{abstract}
  The scheduling of multi-user remote laboratories is modeled as a multimodal function for the proposed optimization algorithm. The hybrid optimization algorithm, hybridization of the Nelder-Mead Simplex algorithm and Non-dominated Sorting Genetic Algorithm (NSGA), is proposed to optimize the timetable problem for the remote laboratories to coordinate shared access. The proposed algorithm utilizes the Simplex algorithm in terms of exploration, and NSGA for sorting local optimum points with consideration of potential areas. The proposed algorithm is applied to difficult nonlinear continuous multimodal functions, and its performance is compared with hybrid Simplex Particle Swarm Optimization, Simplex Genetic Algorithm, and other heuristic algorithms.
\end{abstract}





\maketitle

\section{Introduction}

The laboratory is an essential and supplementary part of learning to bridge the gap between theory and practice. These days, new technologies provide laboratories with a different form of observation, experimentation, and investigation whereby distance learning garnered the attention of researchers. Hence, the remote lab has been designed to perform the ubiquitous platform from faraway places for learners. The most important feature of remote laboratories is to share expensive resources across the world \cite{1-ma2006hands,2-abdulwahed2008beyond}. In this regard, some benefits such as flexible access [3], sharing resources \cite{4-harward2008ilab, 5-lowe2012labshare, 6-richter2011lila}, shared architecture (e.g. Sahara labs \cite{7-lowe2009labshare} and MIT's iLab \cite{4-harward2008ilab}), security of users, data and devices \cite{8-gravier2008state} among many other benefits have been shown to increase attention of users. Regarding the shared platform, user allocations have been a significant issue. Thus, proper scheduling can respond to the demand for having access to the remote laboratory. 

Scheduling problems are formulated as a nonlinear optimization problem. Heuristic optimization algorithms are carried out to address this problem. For example, Simulated Annealing (SA) \cite{8-gravier2008state}, Tabu search \cite{11-hertz1991tabu}, Genetic Algorithm (GA) \cite{12-nuntasen2007application}, Ant Colony Optimization \cite{13-nothegger2012solving}, etc., have been proposed as a methodology to deal with the timetabling problem. Ref \cite{14-shiau2011hybrid} and \cite{15-tassopoulos2012hybrid} solved course scheduling by applying Hybrid Particle Optimization and Hybrid Harmony Search, respectively. High school timetabling has been proposed as one of the hard nonlinear problems because of considering resources and events \cite{16-ahmed2015solving}. Ref \cite{17-akpinar2016hybrid} proposed a hybrid algorithm with a combination of ant colony optimization and Large Neighborhood Search (LNS). As an example, the LNS approach carries out the roulette wheel method based on each wealthy neighborhood to address hybrid algorithms disadvantages \cite{18-li2015iterated,19-sze2016hybridisation}. SA and Particle Swarm Optimization (PSO)\cite{} are combined to schedule different types of trains on a single railway track \cite{20-jamili2012solving}. Further, regarding education setting, university timetable planning has been proposed as constraint-satisfaction problems optimizing by GA \cite{21-deris1999incorporating,zandavi2019state}.

Meta-heuristic optimization algorithms fail to respond the online scheduling problem because not only are they designed to solve most of the optimization problems, but also they do not have enough efficiency in time-consuming \cite{22-yen1998hybrid,23-fan2004hybrid}. On this account, Nelder-Mead simplex has been successfully hybridized with the meta-heuristics to increase the rate of convergence. As an example, Nelder-Mead simplex Particle Swarm Optimization (NM-PSO) \cite{24-fan2006genetic,zandavi2017surface}, Nelder-Mead simplex Genetic Algorithm (NM-GA) \cite{24-fan2006genetic}, Stochastic Dual Simplex Algorithm (SDSA) \cite{zandavi2019stochastic} and Simplex Non-dominated Sorting Genetic Algorithm-II (simplex-NSGA-II) \cite{25-pourtakdoust2016hybrid,26-zandavi2018multidisciplinary} utilize the simplex algorithm as a supplementary part of the main algorithm to improve the exploitation. The main reason is to reach a better compromise between computational efforts and accuracy. 

Developing an efficient optimization algorithm is investigated to make better organization when many requests occupy the shared platform. The proposed algorithm is generated to solve scheduling problems which are modeled as multimodal functions. The proposed algorithm consists of two parts; the simplex approach for increasing exploration and NSGA as finding promising areas. When NSGA sorts local optimum points with consideration of a potential area that probably contains a global minimum, the simplex techniques provide a whole searching domain with a diverse exploration. Hence, NSGA and the simplex part have enough potential to detect promising area and increase the diversity of individuals, respectively.

The organization of this paper is as follows: remote lab timetabling problem is explained in section 2. The Nelder-mead simplex algorithm is detailed in section 3. The non-dominated sorting genetic algorithm is represented in section 4. Section 5 provides the proposed algorithm. Numerical results are made in section 6. The conclusion is drawn in section 7.

\section{Remote lab Timetabling Problem}

The remote lab has been organized as a ubiquitous education technology to expose a set of lab equipment kinds to the world via the internet. Accessing to the appropriate device can play a supplementary role for both education and research activities in a variety of science and engineering disciplines. To enable better usage of available expensive equipment, a remote lab service provider can host multiple users at the same time to amortize the cost of ownership as much as possible. In many situations, allocating the available lab facilities to meet the users' demand is not a straightforward decision. The main constraint is the limited number of available resources versus the possibility of abundant users' request to employ a set of lab resources.
To begin with, handling such shared access among users with different preferences, employing an optimization technique for resource allocation and scheduling strategy to satisfy a set of pre-defined constraints is inevitable. The remote lab timetabling problem involves assigning a set of a limited number of available resources to users' requests to optimize an objective function that reflects unused resources. Also, the constraints to be met are as follows.

\begin{itemize}
    
    \item {Each user can only employ one rig across any requests.}
    
    \item {The amount of time that a user can employ a rig is limited. Such a limitation needs to be determined based on the nature of the experiment as well as the number of other users waiting in queue to access the same rig.}
    
    \item{None of the students can have inactive time (also referred to as idle time) during the possession of a rig.}
\end{itemize}

The primary aim of this timetable problem is to minimize the number of unused rigs of each rig type. Such a minimization leads to increase in the total number of satisfied users (about the constraints mentioned above).  
Notably, the objective function of this optimization system to be minimized can be modeled as follows:

\begin{equation}
\label{eq1}
    f = \sum_{l=1}^{L} \| [X]_{u \times R}(t,l) \times [[C]_{R \times  1}(l)-[P]_{R \times 1}(t,l)] \|
\end{equation}

where $L$ is the number of rig types, and $R$ is defined as a rig set in each type. $C$ and $P$ are the capacity of rig $R$ and the total number of users who are employing such a rig, respectively. $[x]_{u \times R} \in {0,1}$ is a matrix that is set to one if the associated rig $R$ is applied for an experiment in each time-slot $t$ and rig type $l$, otherwise it is equal to zero. 

\begin{figure}[h]
    \centering
    \includegraphics[trim ={1.2cm 14cm 5cm 3cm},clip,scale = 0.6]{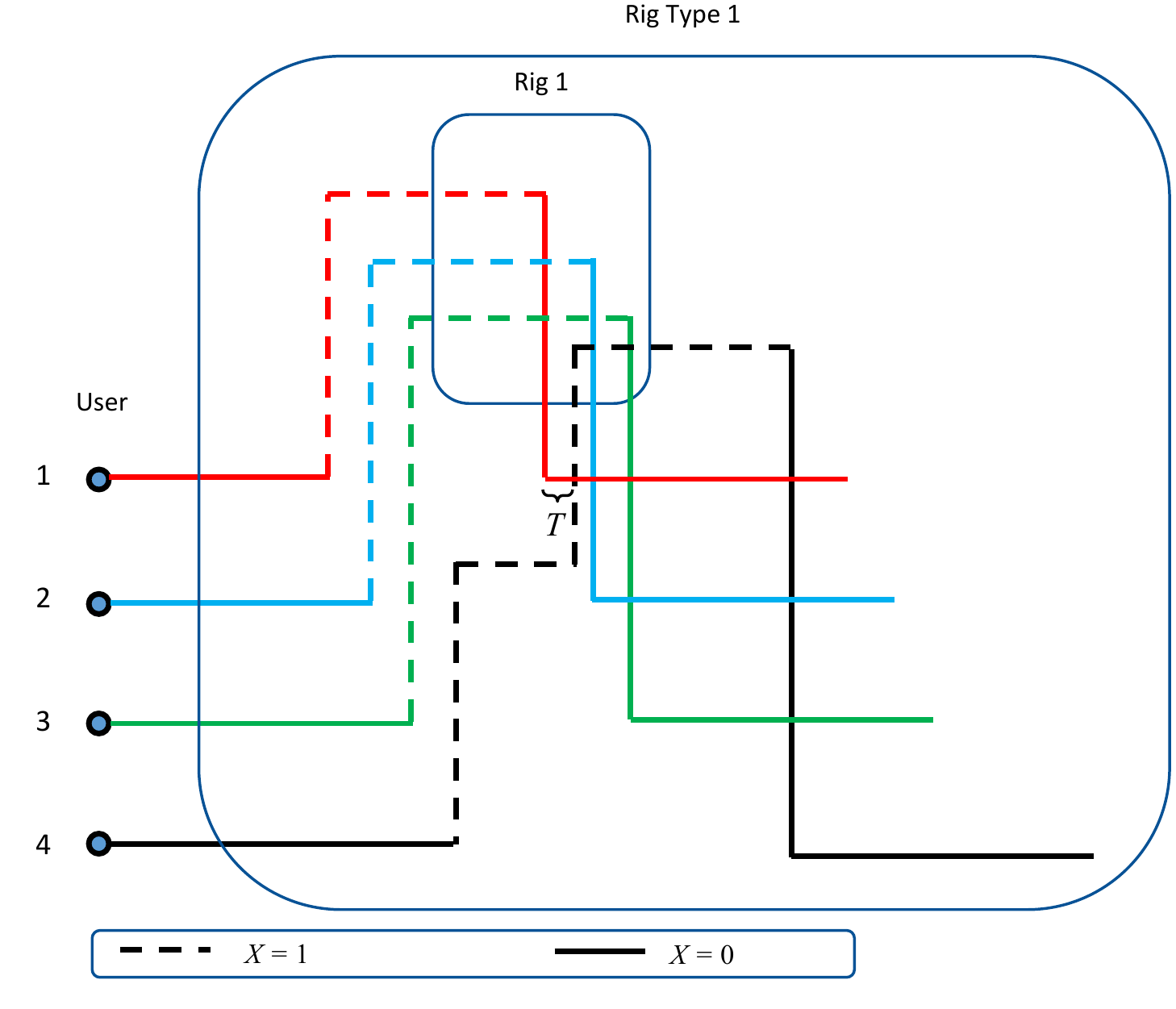}
    \caption{Scheduling strategy}
    \label{fiq1}
\end{figure}

As an example, let us assume that four users want to use a remote lab system consists of $3$ rig types (i.e., $L = 3$). It is assumed that each rig type has one rig (i.e., $R = 1$), while each of them has three capacity levels (i.e., $C = 3$). While each of the first three users uses the first rig type that includes the first rig, then the associated $[X(1,1)](t,l)$ is set to 1. So, the value of $[P(1)](t,l)$ is set to $3$. While the requests of forth user to access any other rig types can be delivered without any delay, any request of this user to access rig $1$ must be queued by the system as other users occupy it. In such a circumstance, $[X(1,1)](t,l)$  remains to one while the value of (\textit{C} - \textit{P}) is equal to $0$. By terminating any experiment of the first three users, rig $1$ is ready to be used by the fourth user. This can be inferred from the above equation by noting that (\textit{C} - \textit{P}) is not equal to zero anymore. Figure \ref{fiq1} depicts a schema for the scheduling process of the case mentioned above.

In Figure \ref{fiq1}, $X = 1$ represents a situation when all users select the Rig $1$ in Rig Type $1$. Also, Rig $1$ is not allocated to the fourth user as others occupy it. When an experiment terminates at the defined procedure, the fourth user can be assigned to use Rig $1$ after elapsing the threshold time ($T= 5$ minutes in this example). 
Thus, it is imperative to use an efficient search method to produce the optimal timetable to satisfy all constraints.

\subsection{Simplex Method}
Nelder-Mead Simplex Algorithm method is classified as a heuristic optimization algorithm and direct research method because the objective function is directly utilized to achieve the best optimal point without derivation. A simplex is a geometrical object produced by $(n+1)$ points $(x_0,...,x_n)$ in $n$-dimension space \cite{27-chelouah2003genetic,28-rao2009engineering,29-nobahari2016simplex}. For example, the triangle is a simplex in two-dimension space. The basic idea in the simplex method is to compare the value of the objective function at the $(n+1)$ vertices of a simplex and move the simplex gradually toward the optimum point through an iterative process. The vertices of a regular simplex (equilateral triangle in two dimensions) of size $\textit{a}$, within $n$-dimensional space, are generated by Eq \ref{eq2} \cite{28-rao2009engineering}.

\begin{equation}
\label{eq2}
    \textbf{x}_{j}=\textbf{x}_{0}+p\textbf{u}_{j}+\sum_{\substack{s = 1 \\ s\neq j }}^{n} q \textbf{u}_s
\end{equation}
where $\textbf{x}_0$ is the initial base point and $\textbf{u}_s$ is the unit vector along the coordinate axis $s$ and also:

\begin{equation}
    \label{eq3}
    p = \frac{a}{n\sqrt{2}}(\sqrt{n+1}+n-1)
\end{equation}

\begin{equation}
    \label{eq4}
    q = \frac{a}{n\sqrt{2}}(\sqrt{n+1}-1)
\end{equation}
 
\begin{figure}[h]
    \centering
    \includegraphics[trim ={3.2cm 17cm 3.5cm 2.5cm},clip,scale = 0.6]{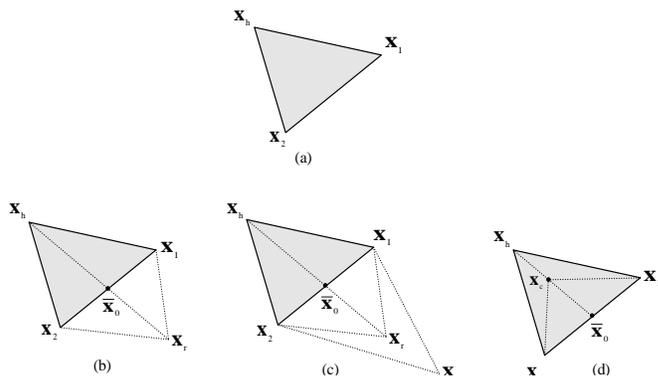}
    \caption{Available moves in the simplex method: (a) initial simplex; (b) reflection; (c) expansion; (d) contraction \cite{29-nobahari2016simplex,25-pourtakdoust2016hybrid}}
    \label{fiq2}
\end{figure}

For reaching the optimal solution, operational tools (reflection, contraction, and expansion) are utilized, deforming the simplex scheme geometrically (see Figure \ref{fiq2}). After each transformation, the current worst vertex is swapped by a better one. Therefore, simplex gradually moves toward to the optimum point.
The reflected, expanded and contracted points are given by: $\textbf{x}_r$, $\textbf{x}_e$ and $\textbf{x}_c$, respectively.

\begin{equation}
    \label{eq5}
    \textbf{x}_r = (1+\alpha)\bar{\textbf{x}}_0 - \alpha \textbf{x}_h \qquad \qquad \alpha > 0
\end{equation}

\begin{equation}
    \label{eq6}
    \textbf{x}_e = \gamma \bar{\textbf{x}}_r + (1-\gamma) \bar{\textbf{x}}_0 \qquad \qquad \gamma > 1
\end{equation}

\begin{equation}
    \label{eq7}
    \textbf{x}_c = \beta \textbf{x}_h + (1-\beta) \bar{\textbf{x}}_0 \qquad \qquad 0 \leq \beta \leq 1
\end{equation}
During these transformations, $\textbf{x}_0$ is the centroid of all vertices $\textbf{x}_0$ except $j = h $; where $h$ is the index of the worst point. The parameters $\alpha$, $\gamma$ and $\beta$ are called reflection, expansion and contraction coefficients, respectively.

Reflection is acting as reflecting the worst vertex, named high, concerning the centroid  . If the reflected point is better than all other points, expansion operates to expand the simplex in the reflection direction; otherwise, if it is at least better than the worst, the algorithm performs the reflection with the new worst point again \cite{28-rao2009engineering}. The contraction is operation because of which the worst point is at least as good as the reflected point.

\subsection{Nondominated Sorting Genetic Algorithm}
NSGA method is a direct optimization algorithm, like simplex. This algorithm is known as a fast approach \cite{30-srinivas1994muiltiobjective}. In this approach, to identify the solution of the first non-dominated front in a population of size $N$, each solution can be compared with every other solution in the population to find out if it is dominated. This process is continued to find all members of the first non-dominated level in the population. This algorithm is utilized Selection, Crossover, and Mutation to find optimal points.

\section{Simplex Nondominated Sorting Genetic Algorithm}
\begin{figure}[h]
    \centering
    \includegraphics[trim ={3.2cm 11.5cm 3.0cm 2.5cm},clip,scale = 0.6]{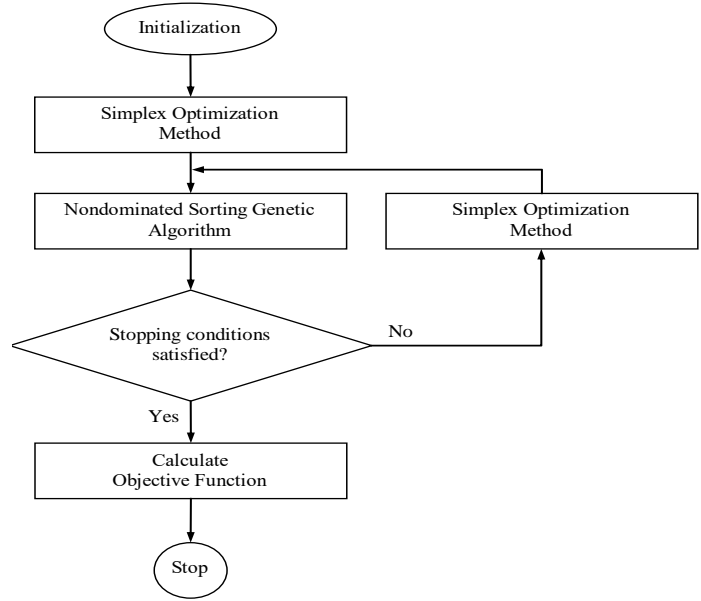}
    \caption{Flowchart of SNSGA algorithm}
    \label{fiq3}
\end{figure}

This algorithm utilizes simplex optimization as exploration and NSGA as exploitation to find the optimal solution. The flowchart of SNSGA is shown in Figure \ref{fiq3}. Also, the parameters of SNSGA are tuned and shown in Table \ref{table1}.

\begin{table}
  \caption{Tuned Parameters of SNSGA}
  \label{table1}
  \begin{tabular}{ccl}
    \toprule
    Parameter & Description & Value\\
    \midrule
    $N_{pop}$ & Population Size & 30\\
    $N_{gen}$ & Maximum Generation & 60\\
    $R_{crs}$ & Crossover Ratio & 1.2\\
    $a_{mut}$ & Scale Parameter & 0.1\\
    $b_{mut}$ & Shrink Parameter & 0.5\\
    $\textit{a}$ & Side of Simplex & 2\\
    $\alpha$ & Reflection Coefficient & 1\\
    $\gamma$ & Expansion Coefficient & 4\\
    $\beta$ & Contraction Coefficient & 0.2\\
    $\textit{i}_{max}$ & Maximum Iteration & 30\\
    
  \bottomrule
\end{tabular}
\end{table}

\subsection{General Setting out of the Algorithm}
SNSGA contains two heuristic algorithms, simplex optimization and NSGA, to reach the optimum global point. In this regard, simplex provides a wide range of variety across search space and acts as exploration; also, NSGA plays a vital role in exploiting the potential area which likely consists of global optimization solutions. Moreover, simplex performs to generate the new population with enough diversity at each generation whereby selection, crossover, and mutation propagate vertices of the simplex. Therefore, simplex operations (reflection, expansion, and contraction) are carried out to deform and move simplex toward likelihood regions of the search space until a maximum number of iterations ($\textit{i}_{max}$) is reached. SNSGA has ten parameters that must be set before the execution of the algorithm. 

\subsection{Initial Population}
In this proposed scheme, the primary parent population is generated randomly with uniform distribution. Not only is a population of each generation sorted based on non-domination scheme, but also each solution is assigned the fitness equal to its non-domination level. Furthermore, recombination based on binary tournament selection is carried out generating a first offspring having the same size as the parent population.

\subsection{Population Update}
In each generation, the population is sorted based on fast non-domination strategy. Simplex part of proposed algorithm organizes the new parent population. Simplex performs to generate the new population with enough diversity, updating population through reflection, expansion and contraction operators for the next generation. Further, any individual that is generated by operational genetic tools is used as vertices of the simplex to achieve new population.

\subsection{Stop Condition}
Simplex part of the proposed algorithm performed in the inner loop terminates at the maximum number of iterations. NSGA part of the algorithm operated in outer loop satisfies termination by the maximum number of generation.
Figure \ref{fiq3} illustrates that the parameters of SNSGA, listed in Table \ref{table1}, must be set before running the algorithm. Then, initial populations that are appropriated for NSGA are produced by simplex method optimization. Therefore, the simplex method helps to increase exploration. Having created a population, NSGA carries out crossover and mutation to find the optimum global point. Afterward, the SNSGA will be stopped if stopping condition satisfies.

\section{Numerical Results}

SNSGA is applied on the benchmarks, listed in \cite{20-jamili2012solving}. According to Ref \cite{20-jamili2012solving}, SNSGA is executed $100$ times to measure the rate of successful minimization, the average of the objective function evaluation numbers, and the average error on the objective function. Once either one of the termination criteria is first reached, the algorithm stops and returns the coordinates of a final point as well as the final optimal objective function value ($FOBJ_{ALG}$ (algorithm)). Analytical minimum objective value ($FOBJ_{ANAL}$) is compared with $FOBJ_{ALG}$, and thus the solution is said to be "successful" if the following inequality holds:

\begin{equation}
    \label{eq8}
    |FOBJ_{ALG}-FOBJ_{ANAL}|<10^{-4}|FOBJ_{INIT}| + 10^{-6}
\end{equation}
where $FOBJ_{INIT}$ is an average of the objective function. 

The average of the objective function evaluation numbers is only accounted for the "successful minimization". The average error is defined as the average of FOBJ deviation between the best successful point and the known global optimum, where only the "successful minimization" achieved by the algorithm. The results of SNGSA tests performed over ten benchmarks are shown in Table \ref{table5}.

\begin{table}
  \caption{Computational Results of SNSGA for the Benchmarks}
  \label{table5}
  \centering
  \begin{tabular}{p{1.2cm}p{2.2cm}p{1.9cm}p{2cm}}
    \toprule
    Test Function & Rate of successful minimization & Average of objective function numbers & Average gap between the best successful point and the known global optimum\\
    \midrule
    RC	& 100 & 109 & $1e-6$\\
    GP  & 100	& 124 & $8e-5$\\
    B2	& 100	& 94 &	$1e-6$\\
    SH	& 100	& 206 &	$5.5e-5$\\
    $R_2$	& 100	& 189 &	$4e-6$\\
    $Z_2$	& 100	& 227 &	$5e-6$\\
    $H_{3,4}$ &	100	& 185 &	$1.35e-4$\\
    $S_{4,5}$	& 98 &	345 & $7e-5$\\
    $R_5$	& 100	& 105 &	$3e-5$\\
    $R_{10}$	& 100 & 148 & $9e-5$\\
    
  \bottomrule
\end{tabular}
\end{table}

The performance of SNGSA is compared with other algorithms such as Continuous Hybrid Algorithm (CHA) \cite{27-chelouah2003genetic}, Enhanced Continuous Tabu Search (ECTS) \cite{31-chelouah2000tabu}, Continuous Genetic Algorithm (CGA) \cite{32-chelouah2000continuous}, Enhanced Simulated Annealing (ESA) \cite{33-siarry1997enhanced}, Continuous Reactive Tabu Search minimum (CRTS min) \cite{34-battiti1996continuous}, Continuous Reactive Tabu Search average (CRTS ave) \cite{34-battiti1996continuous}, Tabu search (TS) \cite{35-cvijovic1995taboo}, INTEROPT \cite{35-cvijovic1995taboo}, Hybrid Nelder-Mead simplex method and Genetic Algorithm (NM-GA) \cite{36-bilbro1991optimization} and Hybrid Nelder-Mead simplex method and Particle Swarm Optimization (NM-PSO) \cite{24-fan2006genetic}. Table \ref{table6} illustrates the average numbers of function evaluation over 100 simulation runs for each benchmark and of the optimization algorithm.

\begin{table}
  \caption{Average Number of Objective Function Evaluation}
  \label{table6}
  \begin{tabular}{lcccccccc}
    \toprule
    Algorithm & $RC$ & $GP$ & $B_2$ & SH & $R_2$ & $Z_2$ & $H_{3,4}$ & $S_{4,5}$\\
    \midrule
    CHA \cite{27-chelouah2003genetic} & 295 & 259 & 132 & 345 & 459	& 215 &	492 &	598\\
    ECTS \cite{31-chelouah2000tabu} & 245 & 231 & 210 &	370 & 480 & 195 & 548 & 825\\
    CGA \cite{32-chelouah2000continuous} & 620 & 410 & 320 & 575 & 960 & 620 & 582	& 610\\
    ESA \cite{38-pradhan2012solving} & - & 783 & - & - & 796 & 15820 & 698 & 1137\\
    CRTS min \cite{39-velazquez2014multi} & 41 & 171 & - & - & - & - & 609 & 664\\
    CRTSave \cite{39-velazquez2014multi} & 38 & 248 & - & - & - & - & 513 &	812\\
    TS \cite{40-hemmatian2014optimization} & 492 & 486 & - & 727 & - & - & 508 & - \\
    INTEROPT \cite{41-shi2015multi} & 4172 & 6375 & - & - & - & - & 1113 & 3700\\
    NM-GA \cite{42-hancer2015multi} & 356 & 422 & 529 & 1009 & 738 & 339 & 688 & 2366\\
    NM-PSO \cite{42-hancer2015multi} & 230 & 304 & 325 & 753 & 440 & 186 & 436 & 850\\
    \hline
    SNSGA & 109 & 124 & 94 & 206 & 189 & 227 & 185 & 345 \\
    
  \bottomrule
\end{tabular}
\end{table}

The numerical results demonstrate that SNSGA has a significant performance improvement in comparison with other hybrid heuristic algorithms. Overall, there is a considerable decrease in the average number of objective function evaluation for almost all benchmarks except for $Z_2$.
Some of the test functions listed in Table \ref{table6} are utilized to describe the performance of the proposed algorithm in finding of optimum global point efficiently. Note that the objective function is normalized. The normalized objective functions are formulated as follow:
\begin{equation}
    \label{eq9}
    NOF = \frac{OF-min(OF)}{max(OF)-min(OF)}
\end{equation}
where NOF is the normalized objective function in each iteration. OF is the real value of the objective function, while max(OF) and min(OF) are maximum and minimum values of the objective functions respectively, throughout the iteration. 

\begin{figure}[h]
    \centering
    \includegraphics[trim ={3.2cm 13.5cm 3.0cm 2.5cm},clip,scale = 0.5]{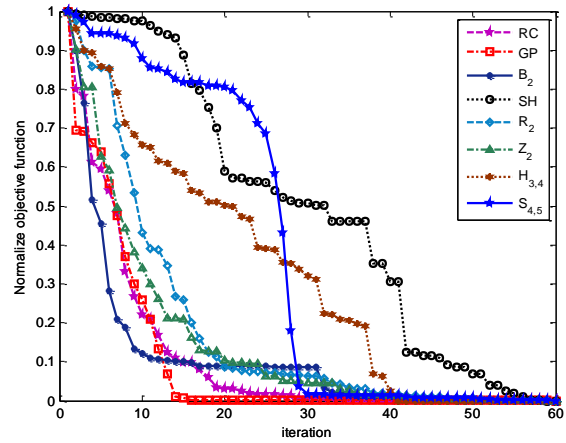}
    \caption{Objective function value of the some benchmarks versus iteration number}
    \label{fiq4}
\end{figure}

Figure \ref{fiq4} represents the convergence performance of SNSGA starting from an initial random point versus the number of iteration. As seen, the SNSGA has precisely reached to global optimums during the iterations. Thus, SNSGA is competitive and even better than other meta-heuristic optimization schemes.

\begin{figure}[h]
    \centering
    \includegraphics[trim ={3.2cm 16cm 3.5cm 2.5cm},clip,scale = 0.7]{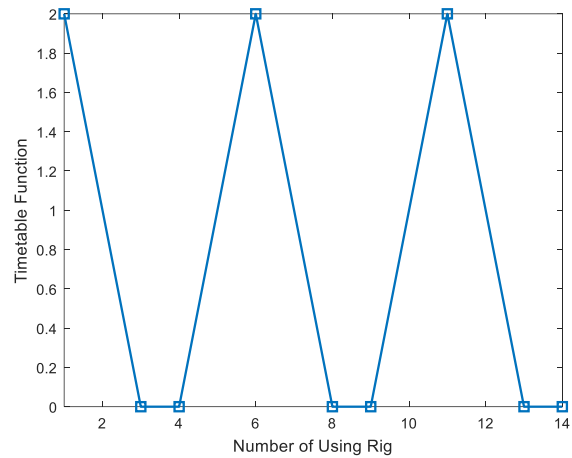}
    \caption{Minimization of the objective function for timetable scheduling mode}
    \label{fiq5}
\end{figure}

SNSGA is utilized to reach the optimum scheme of a scheduling problem for remote labs. Figure \ref{fiq5} shows the results of the optimization for the scheduling problem.
Figure \ref{fiq5} represents the periodical form of timetable problem. This periodical item is completely depended on the number of requests using the specific rig type. When the scheduling process starts with one user, the value of the objective function begins from 2 decreasing gradually in zero based on the request and capacity. Therefore, this trend is periodically occurred to make room for those who cannot use rigs.

\section{Conclusion}
A new hybrid heuristic optimization algorithm was proposed to respond to the demand for having access to the remote laboratories. The proposed hybrid algorithm formulates the scheduling problem as a nonlinear optimization problem. The hybridization of Nelder-Mead simplex and non-dominated sorting genetic algorithm reached a good compromise between time-consuming and accuracy. The numerical results show that the proposed algorithm has a competitive performance in solving scheduling problems.

\bibliographystyle{ACM-Reference-Format}
\bibliography{TimetableSNSGA}


\end{document}